\documentclass[letterpaper, 10 pt, journal, twoside]{ieeetran}
\usepackage{graphicx} 
\usepackage{geometry}
\usepackage{graphics} 
\usepackage{epsfig} 
\usepackage{amsmath} 
\usepackage{amssymb}  
\usepackage{multirow}
\usepackage{url}
\usepackage{siunitx}
\usepackage{color} 

\usepackage[numbers]{natbib}
\usepackage[pageanchor=true,plainpages=false, pdfpagelabels, bookmarks,bookmarksnumbered,hidelinks]{hyperref}
\usepackage{titlesec}
\hypersetup{nolinks=true}
\IEEEoverridecommandlockouts

\title{Towards insect-like distributed proprioception in actuators and appendages for flapping-wing insect-scale aerial robots}
\author{Alexander Hedrick$^{1}$, Arvind Gupta$^{1}$, and Kaushik Jayaram$^{1,2*}$
\thanks{This work is partially funded through grants from the National Science Foundation (Award number 2443869 to K.J.) and this material is based upon work supported by the Air Force Office of Scientific Research under award number FA9550-23-F-0014 in the amount of \$131,900 through the National Defense Science \& Engineering Graduate (NDSEG) Fellowship Program (to A.H.).}
\thanks{$^{1}$Animal Inspired Movement and Robotics Laboratory, Paul M. Rady Department of Mechanical Engineering, University of Colorado Boulder} 
\thanks{$^{2}$Insect Inspired Intelligence Laboratory, Department of Bioengineering, Imperial College London} 
\thanks{$^{*}${For correspondence, \tt\footnotesize k.jayaram@imperial.ac.uk}}%
}

\begin{document}
\maketitle

\begin{abstract}
    Modern flapping-wing insect-scale air vehicles display agility similar to that of their insect counterparts; however, these impressive maneuvers are only possible with off-board sensors like optical tracking cameras. In this manuscript, we introduce two embedded proprioceptive sensors for insect-scale aerial robots: thin film piezoelectric polymers integrated directly into a driving actuator and a pitching hinge which track stroke and pitch angle, respectively. We fabricate the aforementioned size-agnostic mechanically intelligent structures (sensor-actuator, sensor-flexure) using laminate stack fabrication methods. Chirp experiments with our sensors integrated into an insect-size flapping-wing robot show accurate tracking of stroke (RMSE = 0.44$^\circ$) and pitch (RMSE = 2.44$^\circ$) angles in the relevant frequency range. As the first step towards demonstrating the utility of these sensors for enabling numerous onboard autonomy applications, including closed-loop wingbeat control and sensor fusion with existing insect-scale sensor suites for more accurate proprioception and localization, we show one application for each sensor. The proprioceptive hinge enables collision detection, reducing the chance of permanent damage if the robot's wing collides with an object. The proprioceptive actuator enables asynchronous flapping, which is hypothesized to increase adaptability and efficiency in insects and robots alike. A microrobot equipped with our proprioceptive actuator allows us to test these hypotheses with potential for improving flapping aerial robot performance. We foresee proprioceptive sensors having an important role in progressing both the fields of insect-scale aerial robots and robo-physics due to the bio-inspired nature and high integration level of our sensors.
    
\end{abstract}

\begin{figure}[thb!]
    \centering
    \includegraphics[width=\linewidth]{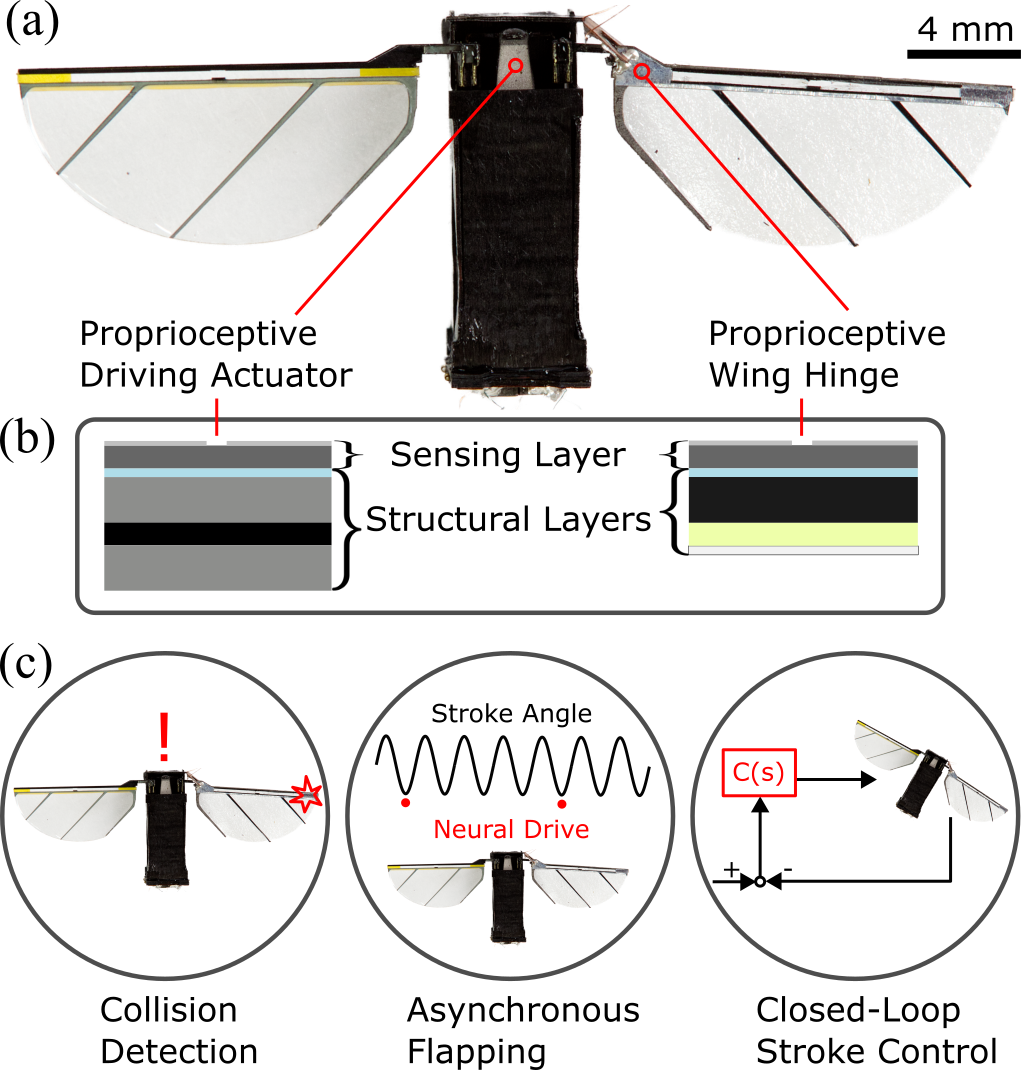}
    \caption{(a) The flapping-wing insect-scale air vehicle with integrated proprioceptive actuator and hinge labeled. (b) Side-view of the stackup for each sensor. (c) Three applications for these sensors: wing collision detection, asynchronous flapping, and closed-loop stroke control. Collision detection is demonstrated in Section \ref{sec:collisions} while asynchronous flapping, a form of closed-loop stroke control, is demonstrated in Section \ref{sec:asynch}.}
    \label{fig:intro}
\end{figure}

\section{Introduction}
\label{sec:intro}

Sub-gram flapping-wing insect-scale aerial vehicles (FWIAVs) have achieved increasingly capable flight, including controlled maneuvering, trajectory tracking, and untethered liftoff  \cite{hsiao_aerobatic_2025, kim_acrobatics_2025, bena_high-performance_2023, chukewad_robofly_2021, jafferis_untethered_2019}. 
These demonstrations, however, commonly rely on external motion capture or other offboard infrastructure for state estimation and control \cite{hsiao_aerobatic_2025, chen_controlled_2019}. Achieving self-contained autonomous operations will require sensing architectures that fit the severe size, weight, and power (SWAP) constraints of insect-scale mechanisms \cite{talwekar_towards_2022, yu_tinysense_2025}. In particular, biological studies \cite{dickinson_wing_1999, salem_flies_2022} reveal that many control-relevant events occur within a wingbeat: changes in stroke amplitude, passive wing rotation, aerodynamic loading, and contact with the environment. These local dynamics cannot be directly measured by conventional body-mounted sensors or inferred reliably from actuator command alone.


This limitation is especially consequential for passively pitching flapping mechanisms \cite{ma_controlled_2013}. In such robots, actuator motion is transmitted through compliant mechanisms to drive wing stroke, while wing pitch emerges from the coupled dynamics of the flexure, wing inertia, transmission, and aerodynamic loading \cite{jafferis_non-linear_2016, chen_experimental_2016}. Consequently, identical actuator commands can produce different wing kinematics when the mechanism experiences changing loads, wear, geometric variation, or contact \cite{lynch_autonomous_2022}. Direct measurement of both \emph{active stroke} and \emph{passive pitch} would therefore provide a local description of wingbeat dynamics that complements body-level inertial or visual sensing \cite{fuller2022gyroscope}. Such measurements could support future feedback control, contact monitoring, and state estimation, but require integration strategies that do not substantially alter the mechanics of the flapping system.


\begin{figure*}[htb!]
    \centering
    \includegraphics[width=\linewidth]{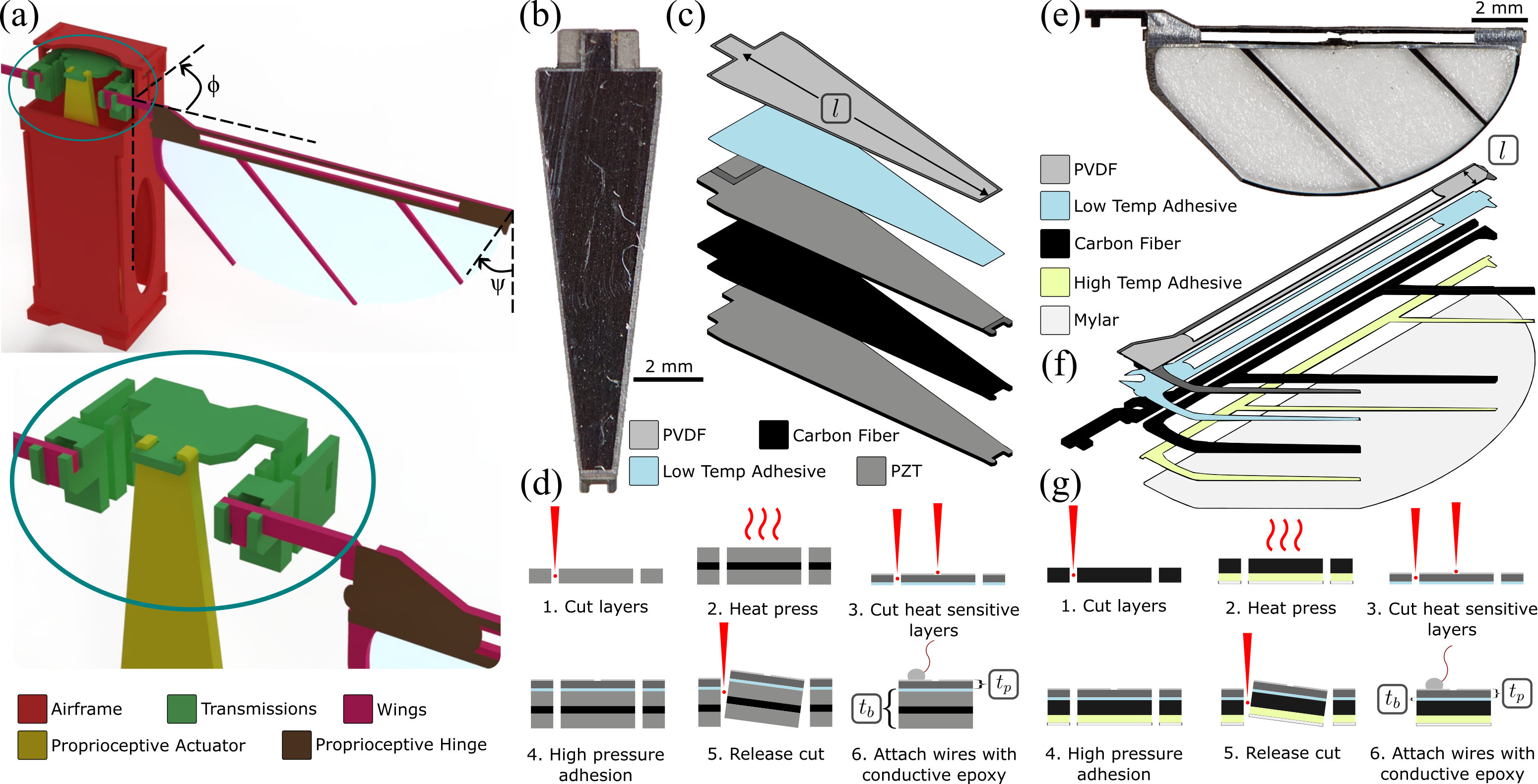}
    \caption{(a) Color-coded CAD model of platform with with stroke ($\phi$) and pitch ($\psi$) angles labeled. (b) Proprioceptive actuator. (c) Color-coded stackup for the proprioceptive actuator showing individual layers. (d) Steps for manufacturing the actuator. (e) Proprioceptive wing. (f) Color-coded stackup for the proprioceptive wing showing individual layers. (g) Steps for manufacturing the wing. }
    \label{fig:design}
\end{figure*}

Prior work has demonstrated promising pathways towards integrated sensing in miniature and flapping robots. Concomitant sensing exploits the electrical response modeling of piezoelectric actuators to estimate their mechanical state \cite{jayaram_concomitant_2018}, while sectioned piezoelectric actuators reserve part of the active structure for sensing \cite{chopra_piezoelectric_2019}. More recently, PVDF layers have enabled dedicated sensing within piezoelectric actuators without sacrificing active actuator area \cite{kabutz_integrated_2025}. Embedded piezoelectric sensing has also been incorporated into flapping-wing systems for deformation, contact, and environmental sensing \cite{hou_scarab_nodate,li_avian-inspired_2024}, and wing strain has been used to infer flight state and support closed-loop control at larger vehicle scales \cite{kim2024wing}. These studies establish the value of embodied sensing but leave an unresolved challenge for insect-scale flapping robots: measuring both the driven and passive components of wingbeat kinematics within the actuator--transmission--wing mechanism without sacrificing performance.

Here, we introduce a distributed proprioceptive sensing architecture for an insect-scale flapping robot (Fig. \ref{fig:intro}). Thin polyvinylidene fluoride (PVDF) films are embedded in (i) a piezoelectric bending actuator to measure stroke and (ii) a passive wing-pitching flexure to measure pitch. 
Such integrated proprioceptive elements represent first steps towards mimicking multifunctional mechanically intelligent biological structures \cite{burden_why_2024} by blurring the lines between \emph{sensor--actuator} and \emph{sensor--flexure}, respectively and enabling onboard autonomy capabilities for FWIAVs. 
Furthermore, proprioception equipped FWIAVs could provide opportunities for robo-physical modeling of similarly-sized organisms, allowing us to test hypotheses about insect intelligence principles \cite{de_croon_insect-inspired_2022}.

The rest of the paper is laid out as follows. 
To enable principled design of distributed mechanosensors, we derived a composite-beam model that applies to both the relatively thick actuator backing and the thin wing-flexure backing, enabling a common sensing framework across these distinct geometries (Sec. \ref{sec:design}).
The propriceptive sensors are fabricated using laminate-stack processes and mechanically integrated at the locations where the strain is coupled to the two principal wingbeat coordinates (Sec. \ref{sec:design}).
We experimentally characterize sensor performance and validate against high-speed-motion ground truth, showing stroke-amplitude tracking with $0.44^\circ$ RMSE and pitch-amplitude tracking with $2.44^\circ$ RMSE over the 110--190~Hz operating range (Sec. \ref{sec:results}).
We conclude with two demonstrations highlighting the utility of proprioceptive information from the embedded mechanosensors: contact-induced deviation sensing in wing-pitch dynamics and delayed-stretch-activation-inspired self-excited flapping using actuator-sensor feedback (Sec. \ref{sec:applications}).

\section{Sensor: Architecture, model, fabrication}
\label{sec:design}
This section describes the flapping mechanism, the locations at which embedded strain sensing encodes wingbeat kinematics, and the fabrication of the resulting sensor--actuator and sensor--flexure structures. The sensing architecture targets two distinct local states: actuator-driven wing stroke and passively generated wing pitch. The platform serves as an integrated testbed for evaluating these measurements under dynamic flapping conditions.



\subsection{Flapping mechanism and sensing locations}

Fig. \ref{fig:design}a shows our insect-scale flapping mechanism used to evaluate the proprioceptive sensors. A piezoelectric bimorph actuator drives two four-bar transmissions through a slider--crank mechanism. The transmissions amplify actuator-tip displacement and convert it into rotational wing stroke ($\phi$). Each wing is attached through a compliant leading-edge flexure that permits passive pitching ($\psi$) under the coupled effects of transmission motion, wing inertia, flexure stiffness, and aerodynamic loading.

Our design improves on manufacturability for sensor integration and experimentation relative to seminal platforms \cite{wood_first_2008}. The transmission and slider--crank are fabricated as a single structure to reduce assembly variation and allow rapid wing replacement during testing. The wing morphology is based on recent designs by Kim et al. \cite{kim_acrobatics_2025} for increased operational longevity. We selected a total flexure width of \SI{4}{\milli\meter} to yield effective pitching dynamics, following prior models of passively rotating flapping wings \cite{chen_experimental_2016}. Our light yet sturdier airframe was optimized to be a mechanically stable base for isolating sensor and wingbeat dynamics. The full platform has a mass of \SI{128}{\milli\gram}.

The two proprioceptive sensors are positioned strategically to best quantify system states ($\phi, \psi$) for flight control. A PVDF film bonded to the bending actuator experiences strain proportional to actuator curvature and, through the transmission kinematics (assumed to be linear), provides an indirect measurement of wing stroke ($\phi$). A second PVDF film bonded to the wing-pitching flexure experiences strain during passive wing rotation and provides a direct measurement of pitch ($\psi$). 
Thus, these measurements are complementary: the actuator sensor reports the actively driven component of the wingbeat, whereas the flexure sensor reports the passive response of the wing mechanism.

\subsection{Sensor design and operational model}

PVDF, a piezoelectric material, generates electric charge when mechanically strained for sensing and vice-versa for actuation \cite{gunter_multilaminate_2025}. 
To measure the bending angle, the PVDF sensing layer is firmly bonded to a backing layer to create a net strain away from the neutral bending axis. 
Both sensors are modeled as two-layer composite beams consisting of a backing ($b$) layer and a sensing ($p$) layer. 
Recent models describing this mechanism typically assume the backing layer to be much thicker than the sensing layer \cite{kabutz_integrated_2025}, which is true for the actuator but not for the flexure. Therefore, we developed the general expression below valid for both sensors.

\begin{table}[!tbh]
    \centering
    \caption{Proprioceptive sensor parameters}
    \begin{tabular}{|c|c|c|c|}
        
        \hline
        Parameter & Symbol & Value & Units \\
        \hline
        \multicolumn{4}{|c|}{PVDF Material Properties} \\
        \hline
        Piezo Stress Constant & $g_{31}$ & 216 & mVm/N \\
        Relative Dielectric Constant & $\epsilon_r$ & 13 & unitless \\
        PVDF Young's Modulus & $E_p$ & 2.8 & GPa \\
        PVDF Thickness & $t_p$ & 12 & $\mu$m \\
        \hline
        \multicolumn{4}{|c|}{Proprioceptive Actuator Parameters} \\
        \hline
        Sensor Area & $A$ & 24.1 & mm$^2$ \\
        Backing Thickness & $t_b$ & 305 & $\mu$m \\
        Length & $l$ & 11.2 & mm \\
        Backing Young's Modulus & $E_b$ & 50 & GPa \\
        Charge Amplifier Gain & $K_{ca}$ & 1.6 & 1/nF \\
        \hline
        \multicolumn{4}{|c|}{Proprioceptive Wing Parameters} \\
        \hline
        Sensor Area & $A$ & 1.2 & mm$^2$ \\
        Backing Thickness & $t_b$ & 5 & $\mu$m \\
        Length & $l$ & 0.3 & mm \\
        Backing Young's Modulus & $E_b$ & 9 & GPa \\
        Charge Amplifier Gain & $K_{ca}$ & 1.7 & 1/nF \\
        \hline
        
    \end{tabular}
    
    \label{tab:parameters}
\end{table}

For a PVDF sensing film operated in the thickness-polarized direction, the generated charge ($Q_3$) is quantified \cite{sirohi_fundamental_2000} as a function of film geometry ($A$ area, $t_p$ thickness), material properties ($g_{31}$ stress constant, $\epsilon_r$ dielectric constant, $E_p$ Youngs modulus), average strain along the length of the sensor ($\varepsilon_1$), and the permittivity of free space ($\epsilon_0$) as:
\begin{equation}\label{eq:charge}
    Q_3 = g_{31}\epsilon_0\epsilon_rE_pA\varepsilon_1
\end{equation}

Assuming Euler-Bernoulli theory for an ideal cantilever beam, a moment applied at the free end results in a linear strain profile. Therefore, the average strain is equivalent to the strain at the centroid of the film. To compute this strain value, we first describe the sensor's neutral axis ($z_{na}$) as:
\begin{equation}\label{eq:zna}
    z_{na} = \frac{E_bt_b\frac{t_b}{2} + E_pt_p(t_b + \frac{t_p}{2})}{E_bt_b + E_pt_p}
\end{equation}
where $E_i$ and $t_i$ are the elastic modulus and thickness for the respective layers ($i= {t,b}$).

Given beam tip displacement ($\delta$), the average strain as a function of the distance ($z$) between the PVDF film centroid ($z_{p,c}$) and the neutral bending axis of the sensor ($z_{na}$) is: 
\begin{eqnarray} \label{eq:strain}
    \varepsilon_1 = -\frac{2z}{l^2}\delta \\
    z = z_{p,c} - z_{na} = tb + \frac{tp}{2} - z_{na} \nonumber
\end{eqnarray}

Combining Equations \ref{eq:zna} and \ref{eq:strain}, we compute strain as:
\begin{equation} \label{eq:strainfinal}
    \varepsilon_1 = -\frac{E_bt_b(t_b + t_p)}{(E_bt_b + E_pt_p)}\frac{\delta}{l^2}
\end{equation}

Under the case that the backing layer is much thicker than the PVDF ($t_b >> t_p$), we recover the equation from \cite{kabutz_integrated_2025}. 
Combining Equations \ref{eq:charge} and \ref{eq:strainfinal}, charge across the sensor as a function of tip displacement is quantified as:
\begin{equation}\label{eq:combined}
    Q(\delta) = -g_{31}\epsilon_0\epsilon_rE_pA\frac{E_bt_b(t_b + t_p)}{(E_bt_b + E_pt_p)}\frac{\delta}{l^2}
\end{equation}

Table \ref{tab:parameters} gives the relevant PVDF material properties along with parameters describing each sensor. 
We use a charge amplifier circuit (Fig. \ref{fig:setup}c) to turn this charge accumulation into voltage for analog measurement. We made a custom multi-channel charge amplifier PCB with components selected for low-noise operation. However, future work could focus on developing a PCB optimized for low SWaP operation, enabling fully onboard sensor operation.
Equation \ref{eq:chargeamp} relates the voltage output (measured) to input charge as a function of sensor tip displacement $\delta$, circuit offset ($V_{HREF} = 1.25$ V), and a tunable gain ($K_{ca}$) that is a function of passive circuit components (Fig. \ref{fig:setup}c):
\begin{equation}\label{eq:chargeamp}
    V_{out}(\delta) = K_{ca}Q(\delta) + V_{HREF}
\end{equation}
\begin{equation*}
    K_{ca} = \left( 1 + \frac{R2}{R3} \right)\frac{1}{C1}
\end{equation*}

\subsection{Proprioceptive actuator manufacturing}
\label{sec:act}


The sensor--actuator consists of two outer PZT-5H layers (PZT-5H, Piezo.com), a carbon-fiber center layer (M55J-RS3C, Toray), and a surface-bonded PVDF sensing layer (Fig. \ref{fig:design}). 
The PZT layers provide bending actuation, while the PVDF layer is electrically isolated from the drive electrodes and used exclusively for strain sensing \cite{kabutz_integrated_2025}. The PVDF layer is placed away from the composite neutral axis to maximize strain during actuator bending.

Fig. \ref{fig:design}b shows a photo of the proprioceptive actuator, and Fig. \ref{fig:design}c shows the stackup. 
Fig. \ref{fig:design}d summarizes the overall manufacturing steps of the actuator, and the details are as follows. 
Fabrication follows a two-stage laminate-stack process \cite{mcdonnell_soft_2026, jayaram_scaling_2020} designed to avoid exposing PVDF to the temperatures above the Curie temperature ($\sim$100$^\circ$ C, \cite{kabutz_integrated_2025}) used to cure the structural actuator layers ($>$175$^\circ$ C for $>$2 hours). First, the PZT-5H and M55J-RS3C carbon-fiber layers are laser patterned, aligned using registration pins, and heat pressed following established high-energy-density piezoelectric actuator fabrication methods \cite{jafferis2021streamlined}. Second, a 12~$\mu$m uniaxially poled PVDF film and 5~$\mu$m pressure-sensitive adhesive (Nitto Inc.) are laser patterned and bonded to the cured structural stack without thermal curing. The PVDF perimeter is rastered during laser processing to ensure electrical isolation of opposing electrode surfaces. Low-fluence femtosecond laser processing is used for the PVDF layer to minimize thermal damage \cite{hedrick_femtosecond_2024}. Adding the sensor layer after the high-temperature structural-lamination steps not only enables integration of piezoelectric sensing without modifying the PZT drive stack but also simplifies the actuator design to 5 stacking layers instead of 13 in a previous work \cite{kabutz_integrated_2025}. Finally, the machined actuators are cleaned in isopropyl alcohol with an ultrasonic cleaner (VEVOR PS-30A). 
This entire process takes $\sim$9 hours and yields 10 actuators. 

The completed actuator has 5 electrical connections: 3 to drive the actuator and 2 to read the sensor, 3 of which can be seen in Fig. \ref{fig:design}b, at the top. To access these signals for external interfaces, we bond a mini-breakout board made out of copper-clad polyimide to the base of the airframe. The finished actuator is press-fit into a slot on the airframe base, and secured mechanically with steel reinforced epoxy (J-B Weld). The actuator pads are connected to the respective copper pads on the breakout board with conductive epoxy (8331D Silver Epoxy, MG Chemicals) and soldered to ultrathin wires (44 AWG, Remington \& 5/48 AWG Litz wire, MWS Wire) for downstream electrical interfacing.

\subsection{Proprioceptive wing manufacturing}
\label{sec:wing}

The sensor--flexure is integrated into the leading-edge wing structure that permits passive pitching (Fig. \ref{fig:design}e). The structural wing stack comprises a 2.5~$\mu$m mylar (Chemplex) membrane, high-temperature acrylic adhesive (FR1500, Dupont), a precured $0^\circ$--$45^\circ$--$0^\circ$ carbon-fiber (M55J-RS3C) laminate, and the same pressure sensitive adhesive and PVDF as the sensor--actuator (Fig. \ref{fig:design}f). The $0^\circ$ fiber direction aligns with the leading edge, while the $45^\circ$ layer aligns with the wing spars. The thin PVDF sensing layer along with the adhesive are bonded across the compliant pitching region such that wing rotation strains the film.
The Mylar layer is patterned on a gel carrier (Gel-Pak 0) to avoid movement during processing. After the structural stack is cured, the PVDF film and pressure-sensitive adhesive are laser patterned and bonded at room temperature using the same low-temperature integration approach as the sensor--actuator. A copper-clad fiberglass (Pulsar Pro FX) clamp and ultrafine wires (44 AWG) provide electrical access to the PVDF electrodes.
This process (Fig. \ref{fig:design}g) takes $\sim$4 hours and yields 4 wings, but the number of wings per batch could be easily increased with minimal increase to batch fabrication time. 



\section{Sensor Performance Characterization}
\label{sec:results}


\begin{figure}[b!]
    \centering
    \includegraphics[width=0.9\linewidth]{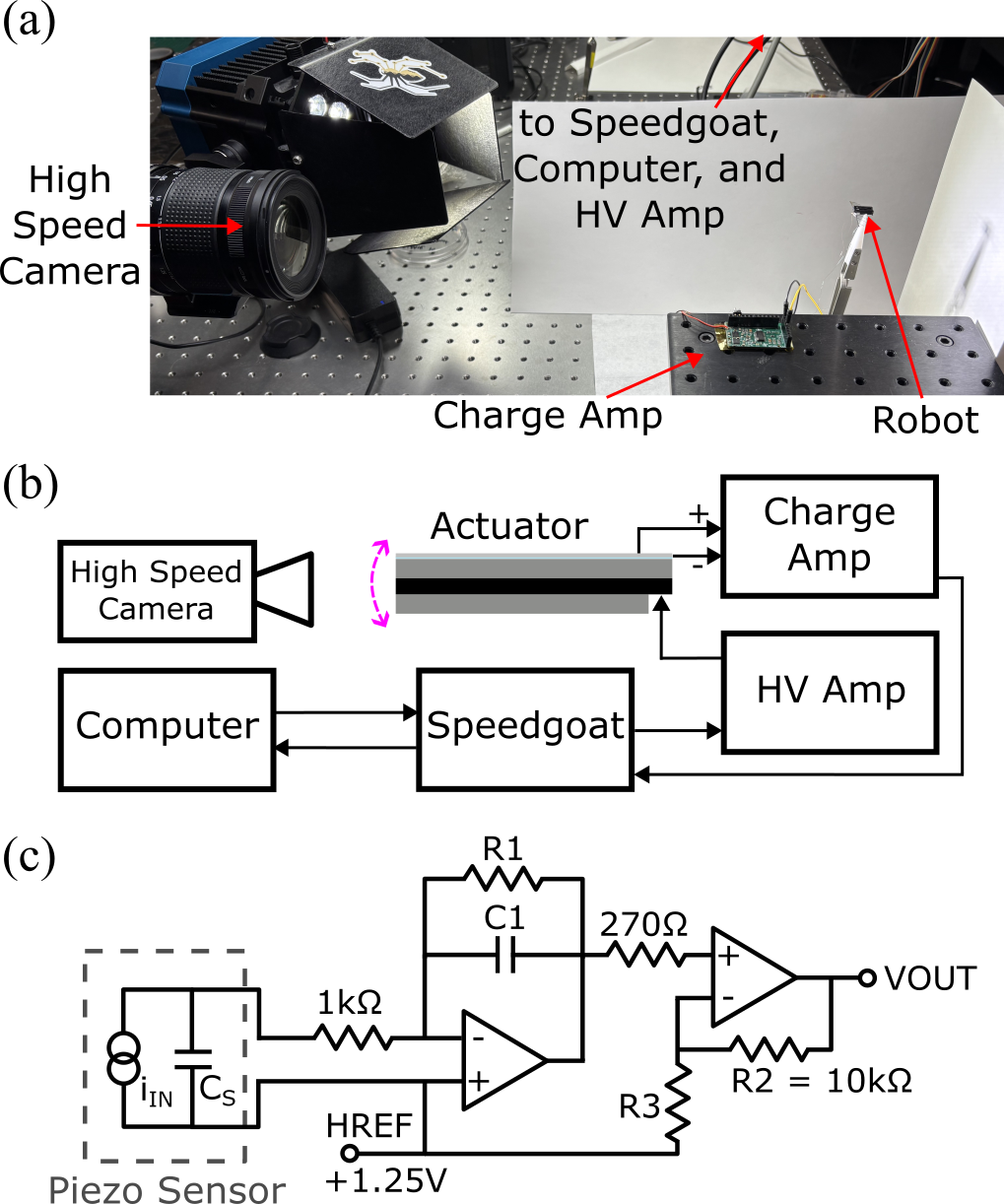}
    \caption{(a) Photo of the setup used for controlling the robot and measuring the output from the proprioceptive sensors. (b) Block diagram of the experimental setup. The real-time Speedgoat computer sends signals to the actuator through a voltage amplifier and receives analog signals from the sensor through a charge amplifier circuit. A high speed camera records the motion of the actuator for post-processing. We use the same setup for tracking the wing motion. (c) Charge amplifier circuit diagram. Equation \ref{eq:chargeamp} describes VOUT as a function of sensor displacement and circuit components.}
    \label{fig:setup}
\end{figure}

We evaluated performance with both proprioceptive sensors integrated into the robot over the relevant dynamic range for typical FWIAVs (60-200 Hz) \cite{jafferis_untethered_2019}. Fig. \ref{fig:setup}a and \ref{fig:setup}b show the setup used for driving the robot while measuring sensor output. A computer communicates with a real-time targeting machine (Speedgoat Inc.) ensuring time-synchronized data acquisition. Inputs to the actuator go through a custom 100x voltage amplifier. Outputs from the sensor go through a custom multi-channel charge amplifier PCB, with the circuit diagram of one charge amplifier shown in Fig. \ref{fig:setup}c. The charge amplifiers and voltage amplifiers are placed on external PCBs for ease of testing but could be easily miniaturized for use on an insect-scale robot. A high-speed camera (Phantom V270) films the robot in action, and revelvant motion is tracked via DLTdv8 \cite{hedrick_software_2008} to quantify ground truth. We determine stroke angle by tracking the actuator tip motion and convert pixel location to stroke angle using the size of each pixel and the robot's transmission ratio. To obtain pitch angle from wing motion, we track three points on the wing (the leading edge and trailing edge on the first spar, and the leading edge on the second spar) and use a custom Python script compute the pitch angle.


\begin{figure}[b!]
    \centering
    \includegraphics[width=\linewidth]{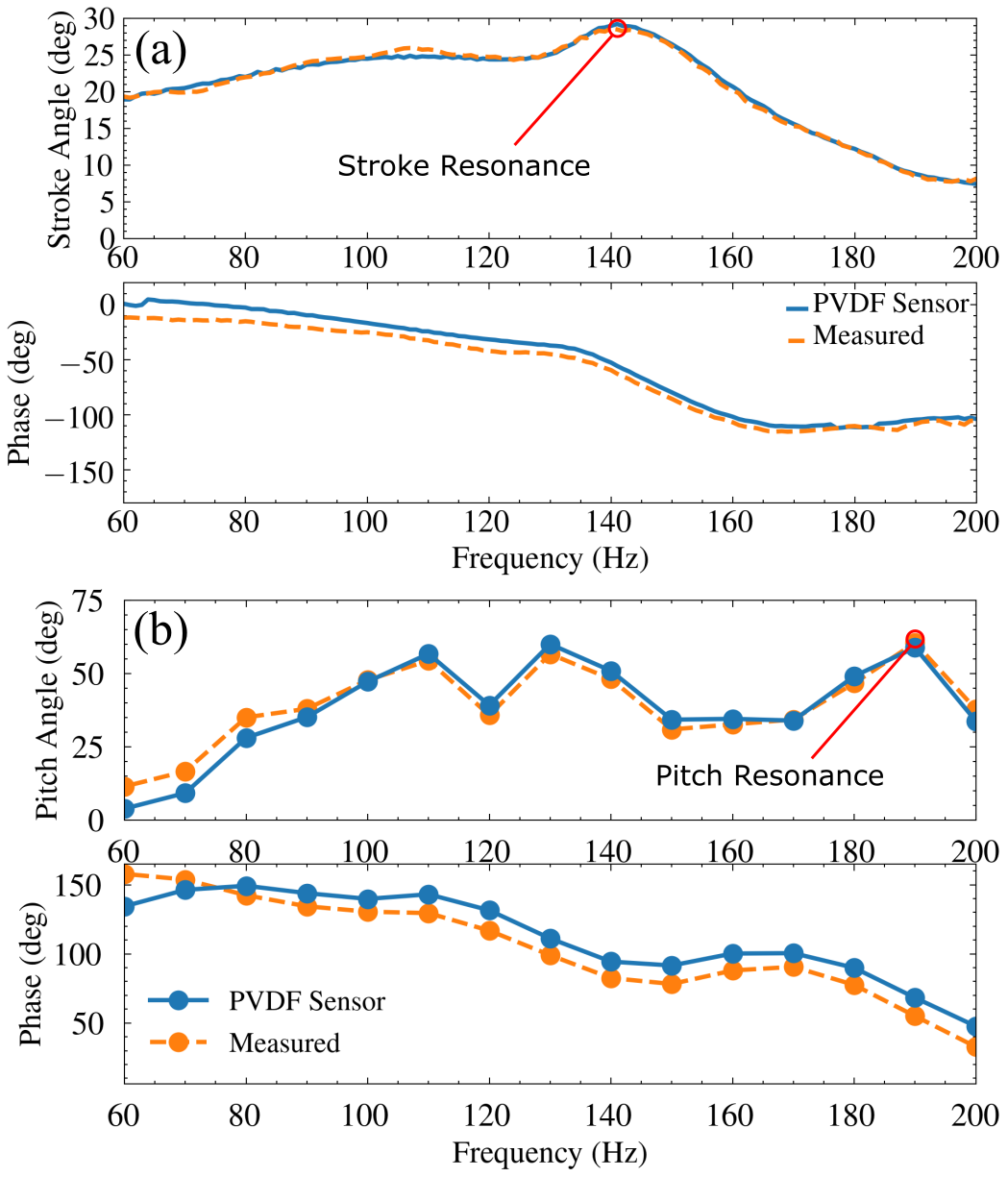}
    \caption{(a) Frequency response for the stroke angle (blue) as calculated from the proprioceptive actuator--sensor data vs. ground-truth (orange). (b) Frequency response for the proprioceptive wing (blue) vs. ground-truth (orange). The magnitude plot shows the pitch amplitude as a function of frequency while the phase plot shows the phase difference between pitch angle and stroke angle.} 
    \label{fig:bode}
\end{figure}

 To quantify the proprioceptive actuator performance, we provided the robot with a chirp signal from 20-200 Hz with $V_{pp} = 200$ V for 1 s. We also performed tests at voltages from 0-175 V in increments of 25 V; however, we only plot the results from the 200 V test for brevity. 
Fig. \ref{fig:bode}a shows the frequency response plot for the proprioceptive actuator. The top plot shows the stroke amplitude of the sensor (blue) vs. ground-truth (orange), and the bottom shows the phase difference between actuator input and stroke angle, both as a function of frequency from 60-200 Hz. We observed the characteristic resonance peak around 140 Hz with a drop in stroke magnitude and increase in phase lag following this point, a common behavior of FWIAV systems \cite{jafferis_non-linear_2016}. We quantified the gain from sensor output (V) to stroke angle (deg) to be 25.5 deg/V. For this gain, we determined the root-mean-square error (RMSE) between the sensor amplitude and ground-truth amplitude to be 0.44$^\circ$.

To quantify flexure sensor performance, we performed a series of single frequency inputs (also $V_{pp} = 200$ V) for 0.2 s to construct a Bode plot from the steady state pitching amplitude.
Fig. \ref{fig:bode}b shows the frequency response plot for the proprioceptive wing. The top plot shows the pitch amplitude of the sensor vs. ground-truth, and the bottom shows the phase difference between the pitch angle and stroke angle. We found the gain from sensor output (V) to pitch angle (deg) to be 108 deg/V. For FWIAVs, the optimal stiffness of the pitching hinge for producing maximal lift gives a pitching resonance higher than the stroke resonance \cite{jafferis_non-linear_2016}. We measure the sensor's performance from 60 Hz to 200 Hz, starting below and ending above both resonance frequencies. The RMSE for the sensor amplitude over the entire frequency range tested is 4.04$^\circ$, calculated by comparing the amplitude at each datapoint in Fig. \ref{fig:bode}b for the sensor vs. measured. In the dynamic range we care about, between the stroke resonance (for this robot $\sim$110 Hz) and the pitch resonance ($\sim$190 Hz), the amplitude RMSE is 2.44$^\circ$. The RMSE for all data averaged across all trials is 8.20$^\circ$; however, most of this error stems from the fact that the sensor consistently lags the actual angle. By correcting for this phase lag, we get an RMSE of 4.44$^\circ$. For the dynamic range we are interested in, 100-190 Hz, the RMSE is 3.93$^\circ$.

As previously mentioned, both sensors consistently lag the ground-truth pitch/stroke angle (RMSE for stroke sensor phase = 9.69$^\circ$ and RMSE for pitch sensor phase = 14.41$^\circ$). Because the lag is consistent, it can be accounted for when designing a controller using these sensors.

\section{Applications}
\label{sec:applications}

The characterization results establish that the embedded sensors track the active stroke and passive pitch quantities during flapping. We next demonstrate the two forms of information that become locally available as a result. First, the flexure sensor reports contact-induced deviations from nominal pitch dynamics. Second, the actuator sensor provides the stroke feedback required to implement delayed-stretch-activation-inspired self-excited flapping. These experiments are proof-of-concept demonstrations of locally embedded sensing; they do not yet constitute closed-loop flight control, quantitative collision classification, or a measurement of energetic or disturbance-rejection benefits, which form exciting future directions.

\subsection{Wing collisions produce measurable pitch deviations}

\label{sec:collisions}

\begin{figure}[htb!]
    \centering
    \includegraphics[width=\linewidth]{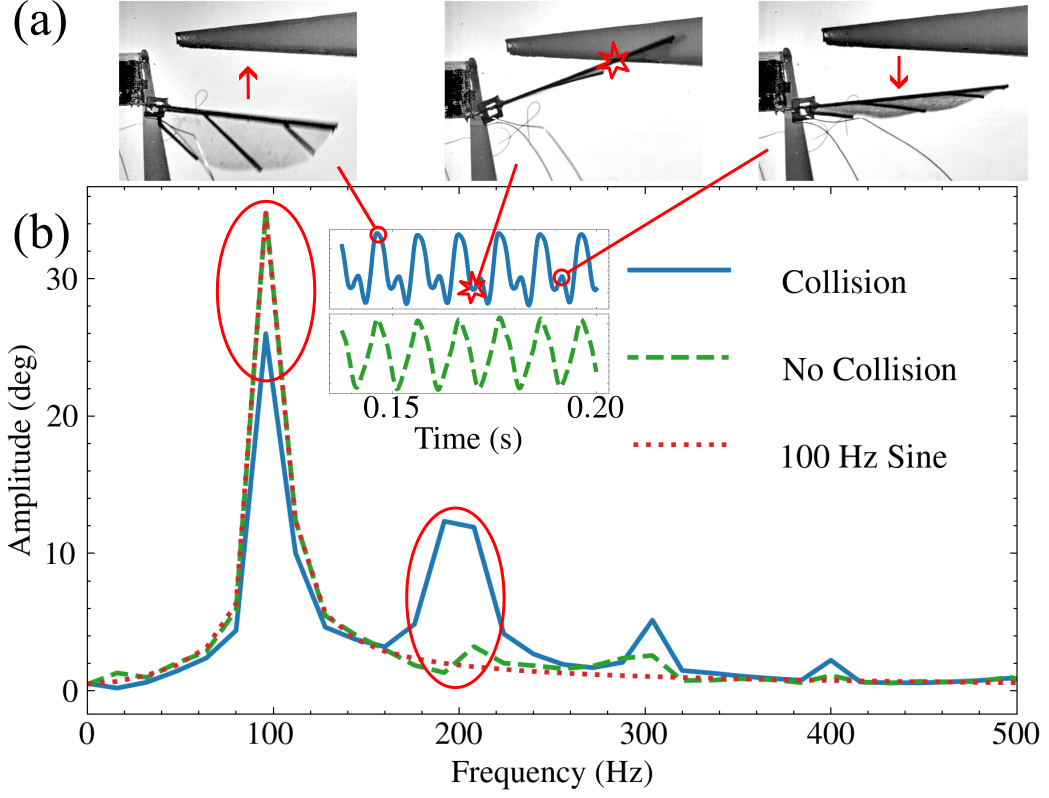}
    \caption{(a) In this experiment we set a solid bar above the wing so the wing collides on every flap and let the system reach steady state. The actuator input is a pure sine wave at a frequency of 100 Hz. Three points are labeled within a wingbeat: before, at, and following the collision. (b) FFT of the collision trial, a no-collision trial at the same input frequency, and a pure 100 Hz sine wave. We include insets of the collision and no-collision trial pitch angles as functions of time. The differences between the collision FFT vs. the no-collision and 100 Hz sine FFTs are circled in red.}
    \label{fig:collisions}
\end{figure}

The next generation of FWIAVs, with onboard sensors, will be able to fly through cluttered and confined spaces. A FWIAV outfitted with only standard sensors (IMU, camera) would not be able to detect wing collisions on a per-wingbeat timescale. To test whether the proprioceptive wing can detect collisions, we run the robot open-loop at 100 Hz and a bias voltage of 150 V with a tweezer set above the wing to collide on every wingbeat. Fig. \ref{fig:collisions}a shows three screenshots from this experiment showing different points in the collision cycle. Fig. \ref{fig:collisions}b shows the fast Fourier transform (FFT) of three signals: sensor-determined pitch angle with collisions (blue), sensor-determined pitch angle \emph{without} collisions (green), and a pure 100 Hz sine wave (red). There are clear deviations from the pure sine wave FFT in the collision FFT that cannot be seen in the no-collision FFT, specifically with a lower amplitude at 100 Hz and a much higher amplitude at 200 Hz. Over the frequency range 0-500 Hz, the RMSE between the collision FFT and the pure sine wave FFT is 3.23$^\circ$ whereas the RMSE between the no-collision FFT and the pure sine wave FFT is only 0.53$^\circ$. The RMSE for the time-data are 15.81$^\circ$ and 3.34$^\circ$, respectively. A controller using this information could set a threshold (such as 1$^\circ$ in our case) for allowable RMSE and change its behavior if the RMSE exceeded the threshold.


\subsection{Embedded stroke feedback enables self-excited flapping}
\label{sec:asynch}

Flapping wing insects can be categorized into having two types of power muscles: synchronous and asynchronous. "Synchronous" insects like moths and butterflies generate wingbeats using periodic signals from the nervous system that match the frequency of the wingbeats. "Asynchronous" insects like flies and bees have specialized power muscles that self-excite to wingbeat frequencies higher than the frequency of nervous system signals used to control the flight, thus "asynchronous" to these signals. Asynchronous actuation is hypothesized to provide advantages to insects such as higher wingbeat frequency, improved efficiency, and adaptability to wing damage \cite{pringle_mechanism_1954, josephson_asynchronous_2000, molloy_kinetics_1987, josephson_efficiency_2001}. Delayed stretch activation (dSA) is the main muscle adaptation enabling self-excited wingbeats. dSA refers to an increase in muscle stress (force) after a time-delay following a strain (stretch) \cite{josephson_asynchronous_2000}.
The dynamics of a dSA system can be modeled as a system of differential equations (Equation \ref{eq:dsa}) \cite{lynch_autonomous_2022, gau_bridging_2023}. 
The first equation describes the stroke dynamics of the nonlinear flapping system where $\phi$ is the stroke angle; $\tau$ is the actuator torque; and $I$, $\Gamma$, and $K$ are wing inertia, aerodynamic damping, and transmission stiffness, respectively. The second equation describes the dynamics of the dSA filter which acts like a second-order low pass filter on the stroke velocity \cite{lynch_autonomous_2022}. The dSA filter constants (Equation \ref{eq:dsaconst}) are functions of three dSA coefficients: $r_3$ is related to force delay, $\kappa$ is related to force decay, and $\mu$ is the system gain.
\begin{equation*}
    I \ddot{\phi} + \Gamma |\dot{\phi}|\dot{\phi} + K \phi = \tau
\end{equation*}
\begin{equation}\label{eq:dsa}
    \ddot{\tau} + a\dot{\tau} + b\tau = -c\dot{\phi}
\end{equation}
\begin{equation} \label{eq:dsaconst}
    a = r_3(1 + \kappa), b = \kappa r_3^2, c = \mu \kappa r_3^2
\end{equation}
A dynamically scaled flapping system has shown adaptive effects and collision detection \cite{lynch_autonomous_2022}; however, the method used to track the velocity of the actuator in that work cannot be scaled down to insect size.


\begin{figure}[htb!]
    \centering
    \includegraphics[width=\linewidth]{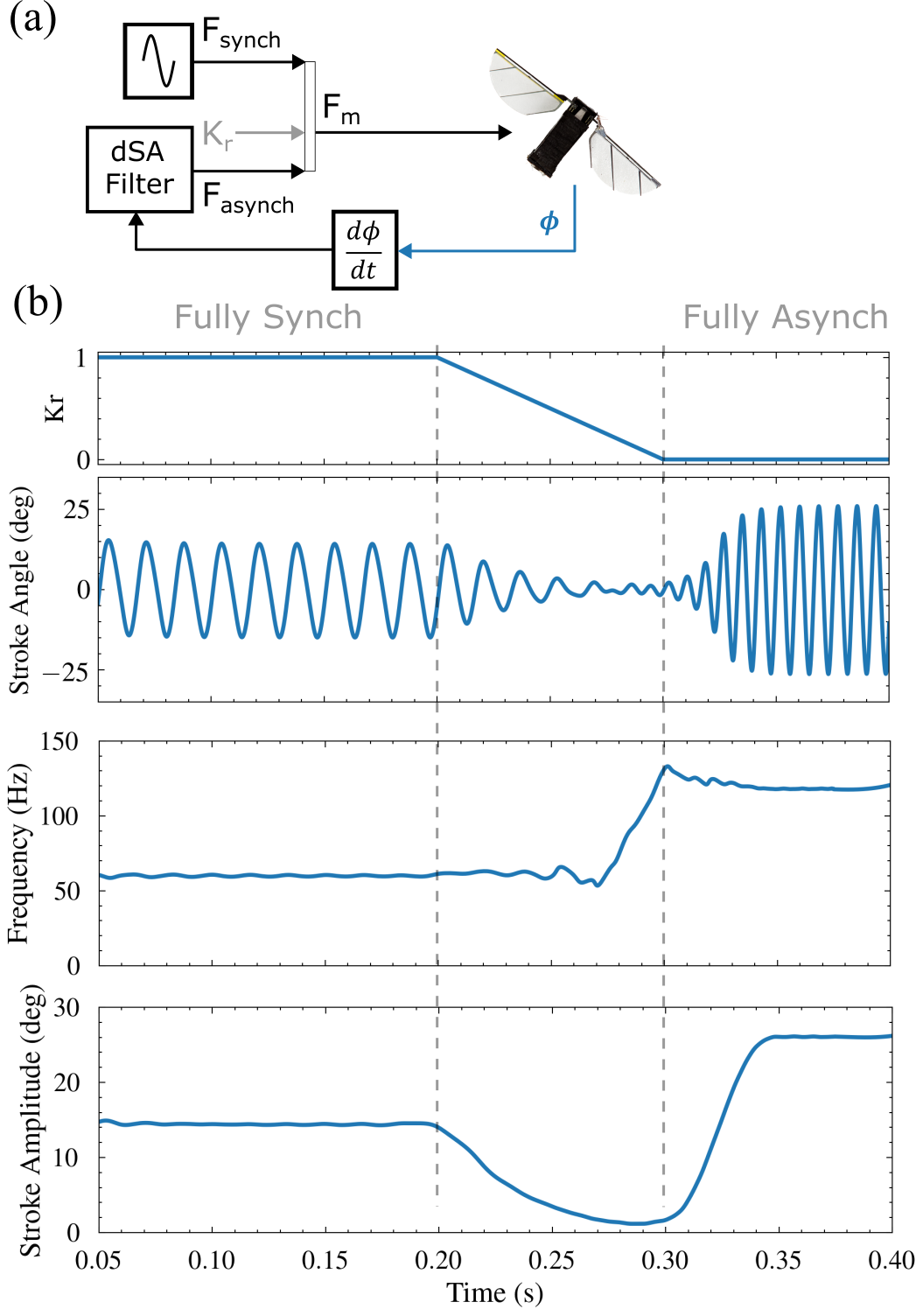}
    \caption{(a) Block diagram for the synchronous to asynchronous transition. Stroke angle is measured with the proprioceptive actuator. The stroke velocity is fed into the delayed stretch activation filter (dSA) which is described by the differential equations \ref{eq:dsa}. The synchronous force is a constant frequency and amplitude sine wave. The parameter $K_r$ determines what fraction of actuator input force ($F_m$) comes from synchronous vs. asynchronous, where $K_r$ = 1 is fully synchronous and $K_r$ = 0 is fully asynchronous. (b) Transition from open-loop synchronous to asynchronous flapping. The first plot shows $K_r$ as a function of time followed by the stroke angle for the trial. The final two plots are the filtered Hilbert transform of the stroke angle plot, showing stroke frequency and amplitude. At the start of the trial, the synchronous frequency is set to a value much lower than the resonance frequency leading to low stroke amplitude. However, as $K_r$ decreases to 0 and the asynchronous dynamics take over, the emergent frequency is slightly higher than the system resonance frequency. The emergent frequency and stroke amplitude are controlled by the three main dSA coefficients: $r_3$, $\kappa$, and $\mu$ which can be seen in the dSA filter coefficients in Equation \ref{eq:dsaconst}.
    }
    \label{fig:asynch}
\end{figure}

In Fig. \ref{fig:asynch}, we show a successful transition from synchronous to asynchronous flapping in our robot using feedback from the proprioceptive actuator as the estimate for stroke angle. Fig. \ref{fig:asynch}a shows the block diagram for the flapping system, and Fig. \ref{fig:asynch}b shows the stroke angle as a function of time as well as the resulting frequency and stroke amplitude, determined using a Hilbert transform of the sensor-determined stroke angle. The Hilbert results are smoothened through a bandpass and Savitzky-Golay filter. This demonstration is an intial step towards onboard generation of self-excited wingbeats in FWIAVs.

\section{Conclusion and future work}

We presented a distributed proprioceptive sensing architecture for an insect-scale flapping mechanism. Thin PVDF films were embedded in a piezoelectric bending actuator and a passive wing-pitching flexure to measure the active and passive components of wingbeat kinematics, respectively. A composite-beam model described the sensing response across the geometrically distinct actuator and flexure structures, and a laminate-compatible fabrication process enabled PVDF integration without subjecting the sensing film to the high-temperature structural curing steps.
Experiments with integrated sensors showed accurate tracking of wing stroke and pitch angle. Finally, we show two applications: collision detection and asynchronous actuation. The sensors introduced in this paper are a step towards fully sensor autonomous FWIAVs. The sensing layers in the proprioceptive structures have negligible weight compared to a robot. Although we used an external PCB to condition sensor signals, the PCB could be miniaturized to fit within insect-scale robot SWaP constraints. We don't expect proprioceptive sensors to replace traditional sensor suites like IMUs and cameras, but rather to complement them. Proprioceptive sensors like the ones in this work are bioinspired solutions that provide information that non-integrated sensors cannot, such as wingbeat dynamics reconstruction, structural health monitoring, and more. Furthermore, our proprioceptive sensors are significantly smaller than existing proprioceptive sensors while retaining similar accuracy; Hou et al. and Li et al. use PVDF for sensing various metrics in a scarab-beetle inspired robot and a bio-hybrid flapping wing robot, respectively \cite{hou_scarab_nodate, li_avian-inspired_2024}. These sensors have a mass on the order of $\sim$0.1-1 g, whereas the sensing elements in our proprioceptive actuator and wing are on the order of $\sim$0.1-1 mg. Thus, our manufacturing technique allows miniaturization of proprioceptive sensors yet to be shown in the literature. Finally, our sensors and the manufacturing technique can readily scaled up, enabling use in any robots that have periodic bending motion, such as larger scale flapping systems and legged robots. 

In this paper, we have shown our proprioceptive wings to be capable of sensing the pitch angle. In addition, because PVDF is a piezoelectric material, we can instead apply voltage across the wing hinge to change the hinge's neutral angle. With the existing unimorph architecture, applying voltage across the hinge would cause the hinge to bend in one direction. This changes the "resting length" of the pitching hinge which can change the dynamics of the flapping system. Thus, with improved PVDF actuator designs \cite{gunter_multilaminate_2025}, future work could involve designing angle-of-attack (AoA) controllers, allowing us to control the attitude of the robot. Dyhr et al. show that the abdomen angle of the moth \textit{Manduca sexta} plays an important role in maintaining stable hovering \cite{dyhr_flexible_2013}. Proprioceptive actuators like the ones developed here can be used to study similar mechanisms.

Finally, we plan to expand upon the applications from Section \ref{sec:applications} in future work, using our platform as a robo-physical model to study biological hypotheses that cannot be answered by studying biological specimen alone. For example, it is not known whether insects with asynchronous muscles are robust to wing damage or adaptive or both. Flies alter their wing kinematics after wing damage to keep the same level of performance as when the wings were undamaged. In other words, the open-loop poles shift to maintain closed-loop performance, and it has been shown that passive changes in dynamics (ie. reduction in wing inertia due to wing damage) are not enough to account for this effect \cite{salem_flies_2022}. However, we are unsure to what extent asynchronous muscles are a factor in flies maintaining performance after wing damage. 
Future work will focus on answering this biological question and similar questions with applications in improving FWIAV performance.

\section*{Acknowledgments}

The authors would like to thank former members of AIM-RL: Dr. Taylor Sharpe, Dr. Heiko Kabutz, Parker McDonnell, and Dr. Hari Krishna Hari Prasad. Also thanks to Dr. Yufeng (Kevin) Chen and Dr. Suhan Kim for help with wing manufacturing.

\bibliographystyle{IEEEtran}
\bibliography{FLAAIR-BIB_2026-07-13}

\end{document}